\documentclass[10pt,twocolumn,letterpaper]{article}

\usepackage{wacv}              

\usepackage{graphicx}
\usepackage{amsmath}
\usepackage{amssymb}
\usepackage{booktabs}

\usepackage{multirow}
\usepackage{booktabs}
\usepackage{graphics}
\usepackage{adjustbox}
\usepackage{subcaption}
\usepackage[font=small,labelfont=bf]{caption}
\usepackage{url}
\usepackage{graphicx}
\usepackage{algpseudocode}
\usepackage{algorithm}
\usepackage{tabularx}
\newcolumntype{Y}{>{\centering\arraybackslash}X}

\usepackage[pagebackref,breaklinks,colorlinks]{hyperref}

\usepackage[capitalize]{cleveref}
\crefname{section}{Sec.}{Secs.}
\Crefname{section}{Section}{Sections}
\Crefname{table}{Table}{Tables}
\crefname{table}{Tab.}{Tabs.}

\def\wacvPaperID{991} 
\def\confName{WACV}
\def\confYear{2024}

\begin{document}

\title{Towards Visual Saliency Explanations of Face Verification}
\author{Yuhang Lu\thanks{Support from XAIface CHIST-ERA-19-XAI-011 and the Swiss National Science Foundation (SNSF) 20CH21\_195532 is acknowledged.}
, Zewei Xu, and Touradj Ebrahimi \\
EPFL, Lausanne, Switzerland \\
{\tt\small firstname.lastname@epfl.ch}}
\maketitle

\begin{abstract}

In the past years, deep convolutional neural networks have been pushing the frontier of face recognition (FR) techniques in both verification and identification scenarios. Despite the high accuracy, they are often criticized for lacking explainability. There has been an increasing demand for understanding the decision-making process of deep face recognition systems. 
Recent studies have investigated the usage of visual saliency maps as an explanation, but they often lack a discussion and analysis in the context of face recognition. This paper concentrates on explainable face verification tasks and conceives a new explanation framework. Firstly, a definition of the saliency-based explanation method is provided, which focuses on the decisions made by the deep FR model. Secondly, 
a new model-agnostic explanation method named CorrRISE is proposed to produce saliency maps, which reveal both the similar and dissimilar regions of any given pair of face images. Then, an evaluation methodology is designed to measure the performance of general visual saliency explanation methods in face verification. Finally, substantial visual and quantitative results have shown that the proposed CorrRISE method demonstrates promising results in comparison with other state-of-the-art explainable face verification approaches. 


\end{abstract}

\section{Introduction}
Recent years have witnessed great advances in face recognition (FR) due to the rapid development of deep learning techniques. Current deep face recognition systems achieve near-perfect performance on well-known public benchmarks and have been widely deployed in several applications, such as access control and surveillance. However, the predictions made by deep learning-based systems tend to be challenging to interpret. The deployment of such biometric systems poses a potential threat to privacy and data protection, resulting in serious public concern. 
To address these issues, it is essential to comprehend the decision-making process of deep face recognition, thereby improving their performance and making them more widely accepted in society. 

This paper focuses on a crucial problem in face recognition, i.e. explainable face verification, specifically by developing insightful explainability tools to interpret the decision-making process of a deep face verification system.
Various saliency map-based algorithms have been proposed as forms of explainable artificial intelligence (XAI) to highlight either the internal CNN layers \cite{zeiler2014visualizing,  olah2017feature, bau2017network} or the important pixels of the input image that are relevant to the model’s decision \cite{simonyan2013deep,binder2016layer,zhou2016learning,zhang2018top,selvaraju2017grad,chattopadhay2018grad,li2018tell}. While many of them have achieved impressive results, they are mainly designed for classification and detection tasks. 


Face verification differs from other vision tasks not only due to the notable difference in the output format, but also the decision-making process, which often involves two images. Explaining an FR model does not mean simply highlighting the critical areas using importance maps but, beyond this, it should also interpret why the given face images are matching or non-matching to the verification system \cite{williford2020explainable}. 
In this context, we first set out and improve the mainstream definition of visual saliency map-based explanations for face verification systems. 
Specifically, an explanation method should reveal the similar regions when the FR system believes the input images are matching and the dissimilar regions if they are non-matching. Then a Correlation-based Randomized Input Sampling for Explanation (CorrRISE) algorithm is proposed, which is model-agnostic and capable of highlighting both the similar and dissimilar regions between any two input face images. 
In addition, the paper proposes a new objective evaluation methodology to compare different state-of-the-art explainable face recognition (XFR) methods in a quantitative manner. 
The contributions of this paper can be summarized as follows:

\begin{itemize}
    \vspace{-1mm}
    \item A model-agnostic explanation method called CorrRISE is proposed to highlight the similarity and dissimilarity regions between any two face images.
    \vspace{-1mm}
    \item A new evaluation methodology is conceived to quantitatively measure the performance of general saliency map-based explanation methods for face verification.
    \item Extensive experiments on multiple face verification scenarios have been carried out and presented, demonstrating the effectiveness of the proposed method.
\end{itemize}

\section{Related Work}
\label{relatedwork}

Visual saliency algorithms have been widely used to explain decision systems in vision tasks that rely on deep learning techniques. A saliency map is essentially an image where each pixel value represents the importance of the corresponding pixel. This map helps identify the significant areas of an input image that contribute to the final output of a ``black-box'' model. In general, there are two types of methods to create such saliency maps. 

The first category of methods involves backpropagating an importance score from the output of the model to the input pixels through the neural network layers. 
Earlier work includes Gradient Backpropagation \cite{simonyan2013deep}, Layer-wise Relevance Propagation~\cite{binder2016layer}, Class Activation Maps (CAM) \cite{zhou2016learning}, and Excitation Backpropagation \cite{zhang2018top}. This type of approach often requires access to the intrinsic architecture or gradient information of the deep model. 
Grad-CAM \cite{selvaraju2016grad} and Grad-CAM++ \cite{chattopadhay2018grad} generalize CAM to be applied to arbitrary convolutional neural networks (CNNs) by weighing the feature activation values with the class-specific gradient information that flows into the final convolution layer. Considering their fame and flexibility in CNNs-based models, this work adapts them to the explainable face verification problem to compare with our proposed method. 

The second category of methods performs random perturbations on the input image, such as adding noise or occlusion, and produces saliency maps by observing the impact on the model's output \cite{ribeiro2016should, fong2017interpretable, dabkowski2017real, petsiuk2018rise}. For example, Ribeiro et al. \cite{ribeiro2016should} proposed an interpretable approximate linear decision model (LIME) in the vicinity of a particular input which analyzes the relation between the input data and the prediction in a perturbation-based forward propagation. The RISE \cite{petsiuk2018rise} algorithm generates random binary masks, applies them to the input image, and then uses the output class probabilities as weights to compute a weighted sum of the masks as a saliency map. 
Our method employs a similar random-masking idea, but differently from RISE, we address a more challenging problem of explainable face verification and propose to incorporate correlation operations into the saliency maps generation process.

While most XAI techniques involving saliency are developed for image classification, there is a growing demand for explanation methods in other image understanding tasks, including object detection \cite{petsiuk2021black} and image similarity search and retrieval \cite{stylianou2019visualizing, dong2019explainability, hu2022x}. In face recognition, earlier work \cite{castanon2018visualizing, williford2020explainable, winter2022demystifying} mainly adapted saliency-based explanation algorithms \cite{zhang2018top, petsiuk2018rise, selvaraju2017grad, selvaraju2017grad} from classification tasks. Alternative research work focuses on face verification models that are explainable by themselves, often referred to as intrinsic explanation methods. For example, Yin et al. \cite{yin2019towards} designed a feature activation diverse loss to encourage learning more interpretable face representations. Lin et al. \cite{lin2021xcos} proposed a learnable module that can be integrated into face recognition models and generate meaningful attention maps. But these methods need to be trained exclusively and thus are not feasible for those deployed verification systems by third parties. Instead, more recent studies \cite{mery2022true, mery2022black} provide ``black-box'' explanations to arbitrary face verification models. The authors introduced several perturbation-based methods to create explainable saliency maps without manipulating or retraining the model and achieved visually promising results. xFace \cite{knoche2023explainable} further improved it by applying more systematic occlusions to inputs and measuring the feature distance deviations. 
However, the evaluation of all the above-mentioned approaches has been based on visualizations, making it difficult to compare to other. This paper provides an objective evaluation methodology and compares our explanation approach with the above-mentioned state-of-the-art model-agnostic methods in a quantitative manner. 





\begin{figure*}[t]
	\centering
	\begin{adjustbox}{width=\textwidth}
    \includegraphics[]{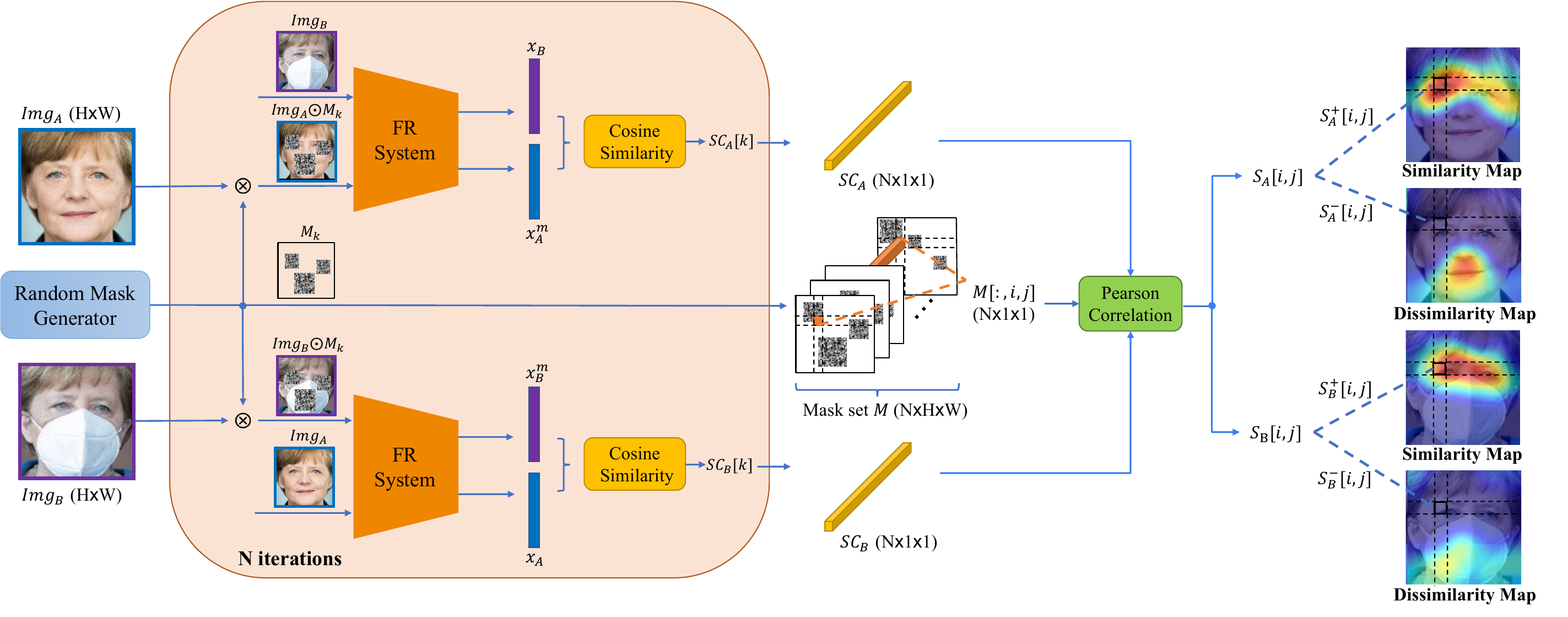}
	\end{adjustbox}
	\caption{Workflow of the proposed CorrRISE explanation method. The similarity and dissimilarity maps are calculated respectively given an arbitrary input face pair. The block in the middle repeats $N$ iterations using different random masks. The output similarity scores and the mask set are fed to the correlation module to calculate similarity and dissimilarity saliency maps in a pixel-wise manner. }
	\label{fig:corrrise}
\end{figure*}



\section{Proposed Method}
\label{method}

\subsection{Definition of Explanation for Face Verification}
In face verification, a deep face-matching model operates by comparing a predefined threshold value with the similarity score between two face images. Ideally, an explanation method should give a visual interpretation of the deep model's decision.

An earlier study \cite{castanon2018visualizing} defined explainability in face verification tasks as a way to visualize the discriminative information of every single face. But practically, the decision-making process of the face recognition model often involves two or more images. The regions that are similar to the FR model may not naturally be the global critical areas. 
Williford et al. \cite{williford2020explainable} have leveraged a face triplet, i.e. probe, mate, and nonmate, to explain the relative importance of facial regions. They defined explainable face matching as a way to highlight the regions of the probe image that maximize the similarity with the mate image and minimize the similarity with the nonmate. However, face recognition systems make decisions independently for each pair of inputs instead of a triplet. 
\cite{mery2022black, mery2022true} improved the definition by directly exploring the relevant parts between two images when the matching is established. But the irrelevant parts between the inputs are not considered, which dominates the decision-making process for non-matching pairs.

This paper summarizes the problem of explainable face verification as follows. 
Given a pair of images feeding into the deep face-matching model, the explanation method should produce the corresponding saliency maps for both input images respectively, which help clearly interpret the prediction results by answering the following questions:

\begin{itemize}
    \item If the face verification system believes the input pair is matching, which regions are \textbf{similar} to the model?
    \item If the face verification system believes the input pair is non-matching, which regions are \textbf{dissimilar} to the model?
\end{itemize}

\subsection{Correlation-based Randomized Input Sampling for Explanation (CorrRISE)}




Most existing explanation methods for face verification interpret the model's prediction with saliency maps that indicate similar regions between any matching images, but they are not eligible for generating meaningful saliency maps for non-matching cases. This section presents a new model-agnostic explanation method called CorrRISE to address the new definition of explainable face verification, see Fig. \ref{fig:corrrise}. 
In principle, CorrRISE generates saliency maps by injecting perturbation and observing the impact on output. Thus, it provides ``black-box'' explanations and can be applied to any FR system without retraining or access to the network. In contrast with other perturbation-based approaches explaining classification models, CorrRISE applies random masks to face images and measures the effect of masked regions on the final similarity scores between two faces rather than a single categorical output. Then, the Pearson correlation between a list of similarity scores and random masks is calculated in a pixel-wise manner to obtain saliency maps. The similar and dissimilar pixels are disentangled from the saliency map according to the correlation coefficients. This is the key innovation of CorrRISE and distinguishes it from previous explainable face verification approaches and prior RISE adaptations by \cite{williford2020explainable,mery2022true}. 
In detail, the CorrRISE algorithm comprises two pivot steps, i.e. mask generation and saliency map generation.

\subsubsection{Mask Generation}
The original RISE algorithm samples binary masks and upsamples them to larger resolutions with bilinear interpolation so that they have values from 0 to 1. Mery \cite{mery2022true} adapted RISE to face verification tasks and generated masks following Gaussian distribution.
We simplify this process here by randomly generating multiple small square patches in various locations of a plain image. 
The values of each patch are not linearly distributed but randomly sampled between $[0,1]$. The parameters are reported in the experimental section. Basically, the mask generation steps are as follows:   

\begin{enumerate}
    \itemsep0.1em 
    \item Initialize the parameters of the mask generator, i.e. the total number of masks $N$, and the number and size of square patches in each mask.
    \item Sample multiple square patches with arbitrary values between $[0,1]$ in random locations of each mask $M_i$ and finally get the mask set $\{M_i, i=1,...,N\}$.
\end{enumerate}

\begin{algorithm}
\caption{}
\label{alg:corrrise}
\begin{algorithmic}
\Procedure{CorrRISE}{} \\
\hspace*{\algorithmicindent}\textbf{Input:} Number of iterations $N$, FR model $f_x$, Similarity function $\texttt{Score()}$, Face images $I_A$ and $I_B$\\
\hspace*{\algorithmicindent}\textbf{Output:} Saliency maps $S^+_A, S^-_A, S^+_B, S^-_B$ 

\State $H, W \gets$ \texttt{Size($I_A$)}
\For{$k=1:N$}
\State $M_k \gets \texttt{RandomMaskGenerator}(H,W)$
\State $x_A, x_B \gets f_x(I_A), f_x(I_B)$
\State $x_A^m, x_B^m \gets f_x(I_A \odot M_k), f_x(I_B \odot M_k)$
\State $SC_{A}[k] \gets \texttt{Score}(x_A^m, x_B)$
\State $SC_{B}[k] \gets \texttt{Score}(x_B^m, x_A)$
\State $M[k,:,:] \gets M_k$
\EndFor

\For{$i=1:H$}
\For{$j=1:W$} \\
\hspace*{\algorithmicindent}\hspace*{\algorithmicindent}\hspace*{\algorithmicindent} $S_A[i,j] \gets $ \texttt{PearsonCorr}($SC_A, M[:,i,j]$) 
\\
\hspace*{\algorithmicindent}\hspace*{\algorithmicindent}\hspace*{\algorithmicindent} $S_B[i,j] \gets $ \texttt{PearsonCorr}($SC_B, M[:,i,j]$)

\EndFor
\EndFor

\State $S^+_A, S^-_A \gets S_A[S_A\ge0], S_A[S_A<0]$
\State $S^+_B, S^-_B \gets S_B[S_B\ge0], S_B[S_B<0]$

\EndProcedure
\end{algorithmic}
\end{algorithm}

\subsubsection{Correlation-based Saliency Map Generation}

Fig. \ref{fig:corrrise} illustrates an overview of the proposed CorrRISE method and Algorithm \ref{alg:corrrise} presents detailed steps. In general, given a pair of images $\{Img_A, Img_B\}$ and a face recognition model $f_x$, the objective is to produce saliency maps that highlight both the similar and dissimilar regions between the two faces. First, CorrRISE leverages a mask generator to randomly produce $N$ masks $M=\{M_i, i=1,...,N\}$. Each of the masks $M_i$ will be multiplied with the input images respectively, e.g. $img_A$. The masked $img_A\odot M_i$ and unmasked $img_B$ are fed into the face recognition model $f_x$ to capture the deep face representation $\{x^m_A, x_B\}$. The cosine similarity score $SC_i$ between the deep features is then calculated. After iterating all the $N$ masks, the list of scores $SC=\{SC_i, i=1,...,N\}$ corresponding to the mask list is recorded. Subsequently, Pearson correlation is performed between $SC$ and $M$ in a pixel-wise manner to obtain the final saliency map $S_A$ for $img_A$. The location of positive correlation coefficients represents the regions on $img_A$ that are similar to $img_B$, while the location of negative coefficients represents the dissimilar regions. 
The same procedure is applied to $img_B$ to get the saliency map $S_B$. We perform these saliency map generation procedures separately for two images, see Fig. \ref{fig:corrrise}, because a face-matching system will verify two irrelevant but masked faces as a matching pair, which interferes generation process.


\section{Evaluation Methodology}
\label{chapter:eval}
Despite the promising development of explainability methods in vision tasks, the importance of rigorous objective evaluation methodologies has long been overlooked. 
In particular, only a few metrics have been designed for visual saliency explanation tools. In the past, human evaluation was the predominant way to assess model explainability \cite{herman2017promise, zhang2018top}. Petsiuk et al. \cite{petsiuk2018rise} proposed two automatic objective metrics, ``Deletion'' and ``Insertion'', for explainable image classification task, which measure the change in output classification probability after removing or adding salient pixels from/to the input image. \cite{hu2022x} adapted these metrics to image retrieval tasks. In face verification, \cite{castanon2018visualizing} quantified the visualized discriminative features by playing a ”hiding game”, which iteratively obscures the least important pixels in the image sorted according to a produced attention map. But it is not precise enough to differentiate high-performing explanation methods \cite{xu2023discriminative}. 

This paper proposes new ``Deletion'' and ``Insertion'' metrics to better fit explanation methods for face verification. Specifically, these metrics measure the change in the verification accuracy after modifying the input image according to the importance map generated by the explanation method. The intuition behind the proposed metrics is that the explanation saliency map is expected to precisely highlight the most important regions of two faces with the smallest number of pixels, based on which the FR model can make final decisions. The faster the overall accuracy drops/rises after removing/adding salient pixels, the more accurate the produced saliency map.

\begin{algorithm}
\caption{Deletion and Insertion Metric}
\label{alg:metrics}
\begin{algorithmic}
\Procedure{Evaluation Metric}{} \\
\hspace*{\algorithmicindent}\textbf{Input:} FR model $f_x$, FR evaluator $\texttt{eval()}$, testing dataset $D$, saliency maps $S$, number of steps $n$\\
\hspace*{\algorithmicindent}\textbf{Output:} deletion score $d_1$, insertion score $d_2$
\State $step \gets 1/n \times 100$
\State $S \gets \texttt{Sorting}(S)$ \algorithmiccomment{sort in descending order}
\For{$i=1:n$}
\State $p \gets i/n\times100$
\State Mask the first $p\%$ pixels of $D$ to get $D^\prime_{1}$
\State Insert the first $p\%$ pixels to plain images to get $D^\prime_{2}$
\State $acc_{1i} \gets \texttt{eval}(f_x, D^\prime_{1})$
\State $acc_{2i} \gets \texttt{eval}(f_x, D^{\prime}_{2})$
\EndFor
\State $d_1 \gets \texttt{AreaUnderCurve}(acc_{1i}$ vs. $i/n,\text{ }i=1:n)$
\State $d_2 \gets \texttt{AreaUnderCurve}(acc_{2i}$ vs. $i/n,\text{ }i=1:n)$
\EndProcedure
\end{algorithmic}
\end{algorithm}

Algorithm \ref{alg:metrics} describes the procedure for calculating the proposed metrics. 
In general, the ``Deletion'' metric executes the following steps. First, the generated saliency map for each input image is sorted according to the importance value. Then, $n$ threshold values are evenly sampled from $[0,1]$, i.e. $p_k = k/n\times 100\%$ where $k\in \{1, \dots, n\}$, and serve as different percentages of modified pixels in the input image. Subsequently, a verification task using a dataset $D$ is performed iteratively as an auxiliary task in the overall evaluation methodology. In each iteration, $p_k$ percent of the most salient pixels are masked from the input image of the entire testing dataset. The accuracy of the face recognition model is then measured on the modified dataset. Finally, the ``Deletion'' metric is defined as the AUC score of the Accuracy vs. Percentage of masked pixels curve. The lower the metric, the more accurate the evaluated saliency map. 
The ``Insertion'' metric takes a complementary approach but performs similar procedures. In each iteration, $p_k$ percent of the most important pixels are inserted into a plain image. Ideally, a higher ``Insertion'' score refers to a better explanation of the saliency map. 


\section{Experimental Results}
\label{results}

\subsection{Implementation Details}
\subsubsection{Explanation Method Setup}
\label{label:implementation}
The proposed CorrRISE explanation method does not require any training or access to the internal architecture of the face recognition model. During the explanation process, the number of generated masks, i.e. number of iterations, is set to 500 by default. For each mask, there are 10 patches and the size of each patch is 30$\times$30 pixels.

As a comparison, two state-of-the-art XFR methods MinPlus \cite{mery2022true}, and xFace \cite{knoche2023explainable} have been tested based on their official open-source code. 
Meanwhile, several XAI methods \cite{selvaraju2017grad,chattopadhay2018grad,ribeiro2016should,petsiuk2018rise} have been adapted and tested. For Grad-CAM \cite{selvaraju2017grad} and Grad-CAM++ \cite{chattopadhay2018grad}, instead of backpropagating the gradients of class-wise posterior probability to activation layers, we adapt them by performing backpropagation for gradients of similarity scores between two input images. Moreover, we utilize the third-party adaptation from authors of \cite{mery2022true} for LIME \cite{ribeiro2016should} and RISE \cite{petsiuk2018rise}.

\subsubsection{Face Recognition Model Setup}
This paper adopts three different state-of-the-art face recognition models for experiments to show the validity of our proposed model-agnostic method, namely ArcFace \cite{deng2019arcface}, MagFace \cite{meng2021magface}, and AdaFace \cite{kim2022adaface}. All the models share the same ResNet-50 \cite{he2016deep} feature extractor and are trained on the same dataset by running the publicly available code. Except for the cross-model analysis, all the experiments are conducted with the ArcFace model.



\subsubsection{Dataset}
The face recognition models are trained with the MS1Mv2 dataset \cite{deng2019arcface}. For evaluation, this paper first selects samples from different datasets, namely LFW \cite{huang2008labeled}, CPLFW \cite{zheng2018cross}, LFR \cite{elharrouss2020lfr}, and Webface-Occ \cite{huang2021face} for visual presentation in various verification scenarios.
In the proposed objective evaluation methodology, LFW, CPLFW, and Webface-Occ datasets are used for quantitative evaluation. 
All the images are cropped and resized to 112x112 pixels.






\begin{figure}[t]
	\centering
	\begin{adjustbox}{width=0.8\linewidth}
    \includegraphics[]{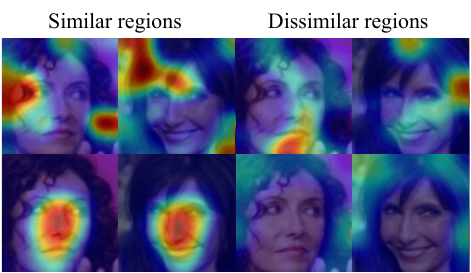}
	\end{adjustbox}
\caption{Sanity check for the CorrRISE explanation method. The first row is the explanation heatmap for a deep model with randomized parameters, while the second row is for a normal face recognition model. The importance increases from blue to red color.}
\label{fig:sanity}
\end{figure}


\begin{figure*}[t]
	\centering
	\begin{adjustbox}{width=0.8\textwidth}
    \includegraphics[]{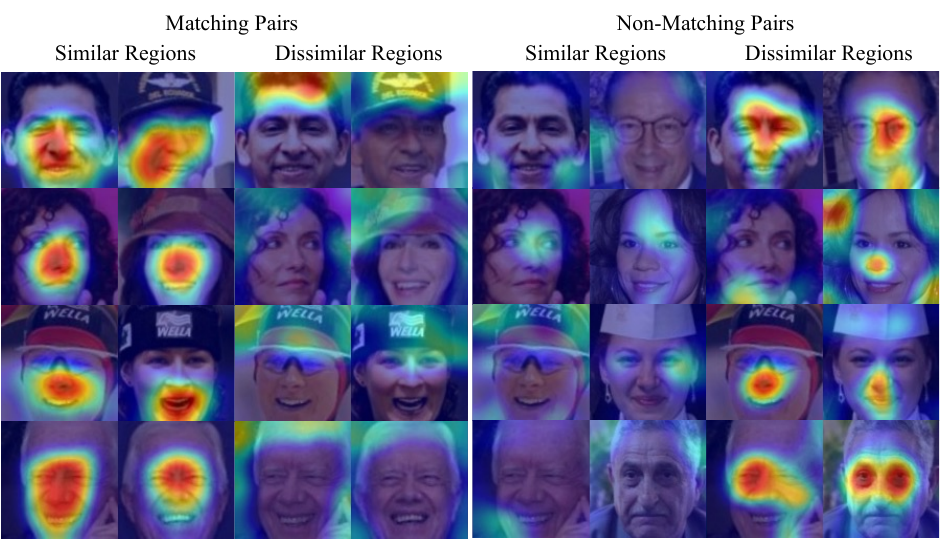}
	\end{adjustbox}
\caption{Visual explanation results from CorrRISE for both matching and non-matching face pairs in standard face verification scenario. The produced saliency maps explain why the verification model makes correct predictions on all face pairs. The saliency value increases from blue to red color. }
\vspace{-2mm}
\label{fig:verification}
\end{figure*}

\subsection{Sanity Check}

A recent study \cite{adebayo2018sanity} has raised doubts about the reliability of visual saliency methods that the produced explanation heatmaps can be independent of the deep model or the input data. They introduced a model parameter randomization test for a sanity check. 
In the context of face verification, an explanation method may provide visually compelling heatmaps by directly emphasizing the center of the faces without interacting with the deep FR model. Therefore, this paper employs a similar sanity check to validate the effectiveness of the proposed method. Specifically, the parameters of the ResNet-50 backbone network are randomly initialized and then the CorrRISE algorithm is applied to the randomized model. 
The first row of Fig. \ref{fig:sanity} shows that the CorrRISE algorithm will generate nonsense saliency maps when it tries to explain a face recognition model with fake parameters. It proves that the proposed explanation method relies on a feasible recognition model and is capable of producing meaningful interpretations.


\subsection{Visual Explanation Results}

This section presents the visual results of the saliency maps generated by our proposed CorrRISE algorithm. For easier demonstration and fair comparison, all the experiments here are conducted on the ArcFace model. First, the explanation ability of CorrRISE is tested on the standard face verification scenario with samples randomly selected from LFW and WebFace-Occ datasets respectively. 
Then, the behavior of the ArcFace model in several challenging scenarios is analyzed and explained, notably some failure cases due to very similar subjects or significant head pose variations. Moreover, a visual comparison with other explanation approaches is presented.


\begin{figure}[t]
	\centering
	\begin{adjustbox}{width=0.8\linewidth}
    \includegraphics[]{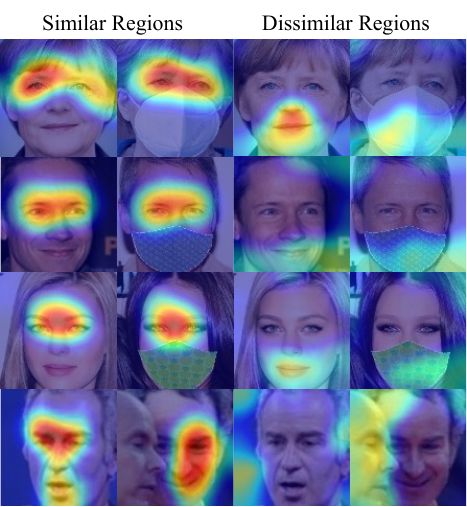}
	\end{adjustbox}
\caption{Saliency map explanations for the predictions of the FR model on partially-occluded faces. The masked regions are accurately identified as dissimilar regions. The saliency value increases from blue to red color.}
\vspace{-2mm}
\label{fig:occlusion}
\end{figure}


\begin{figure*}[t]
	\centering
	\begin{adjustbox}{width=0.8\textwidth}
    \includegraphics[]{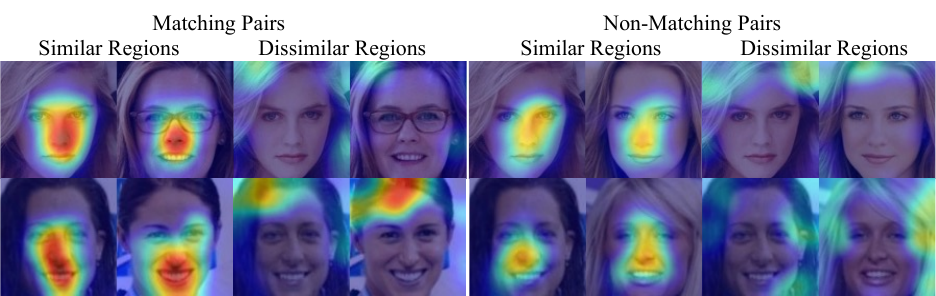}
	\end{adjustbox}
\caption{Saliency map explanation that interprets why the FR model mistakenly matches the two non-matching pairs (right). The saliency value increases from blue to red color.}
\vspace{-1mm}
\label{fig:fail}
\end{figure*}


\begin{figure}[t]
	\centering
	\begin{adjustbox}{width=0.8\linewidth}
    \includegraphics[]{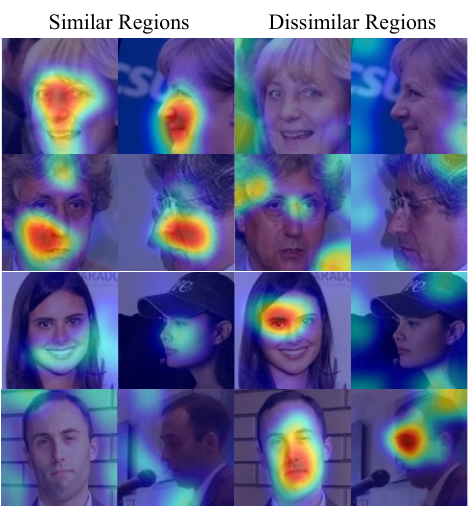}
	\end{adjustbox}
\caption{Saliency map explanations for the predictions of the FR model on matching but ill-posed faces. The FR model makes correct predictions in the first two examples and fails in the rest. The saliency value increases from blue to red color.}
\vspace{-1mm}
\label{fig:pose}
\end{figure}

\subsubsection{Standard Verification Scenario}

In face verification, a deep FR model predicts whether a pair of face images belong to the same identity. Fig. \ref{fig:verification} shows the visual explanations for the model's decision on four matching and four non-matching pairs of images taken from the LFW dataset. Notably, here the deep model makes correct predictions on all eight pairs.

As a result, the saliency maps produced by CorrRISE properly highlight regions that the FR model believes are very similar between the matching pairs. The general salient region focuses on eyes, noses, and mouth, but it also varies from person to person. For example, the FR model relies more on the cheek when comparing the matching pair in the first row while more on the open mouth for the pair in the third row. It is also notable that the dissimilar regions often concentrate on irrelevant backgrounds and unexpected occlusions, such as hats. As for the non-matching pairs, CorrRISE tries to localize the similar areas between the non-matching faces but with rather low salient values. In contrast, it produces saliency maps that clearly spotlight the most dissimilar regions in their faces, indicating very low similarity between them. 

To further show the effectiveness of the proposed explanation algorithm, an additional test has been performed with partially occluded faces, taken from WebFace-Occ. In this experiment, the face recognition model is also capable of verifying occluded faces despite slightly lower similarity scores. 
As presented in Fig. \ref{fig:occlusion}, the CorrRISE algorithm precisely localizes the non-occluded regions which the FR model leverages for correct predictions, whilst it also highlights the occlusions, such as masks and even another person's face, as dissimilar regions.

\subsubsection{Challenging Scenarios}
Face verification systems can suffer from various challenging situations in daily usage, such as very similar identities or head pose changes. 
It is important to provide reliable explanations for the system's behavior in specific scenarios. 

Fig. \ref{fig:fail} presents two examples of similar identity scenarios in a format of triplets. 
The deep model has mistakenly verified both the matching (left) and non-matching pairs (right) as faces from the same subject. This wrong decision is explained by the saliency maps produced by CorrRISE. They show that, although with lower salient values than the matching pairs, the deep model believes the nose and mouth regions of the two non-matching images are close enough to be classified as the same person. Meanwhile, there is no significant dissimilar region between them. 

Head pose variation is another well-known challenge for face verification systems. Fig. \ref{fig:pose} shows four examples, where the deep model correctly recognizes the first two while failing at the last two. The saliency maps corresponding to the first two examples show that the deep model manages to localize similar regions with high salient values even on the sided faces. 
In contrast, it fails to find enough similarities in the last two examples and makes false predictions due to lacking sufficient information. For instance, the dissimilar saliency map spotlights the left eye of the front face in the third example, which corresponds to the missing parts in the sided face.



\begin{figure*}[t]
	\centering
	\begin{adjustbox}{width=0.85\linewidth}
    \includegraphics[]{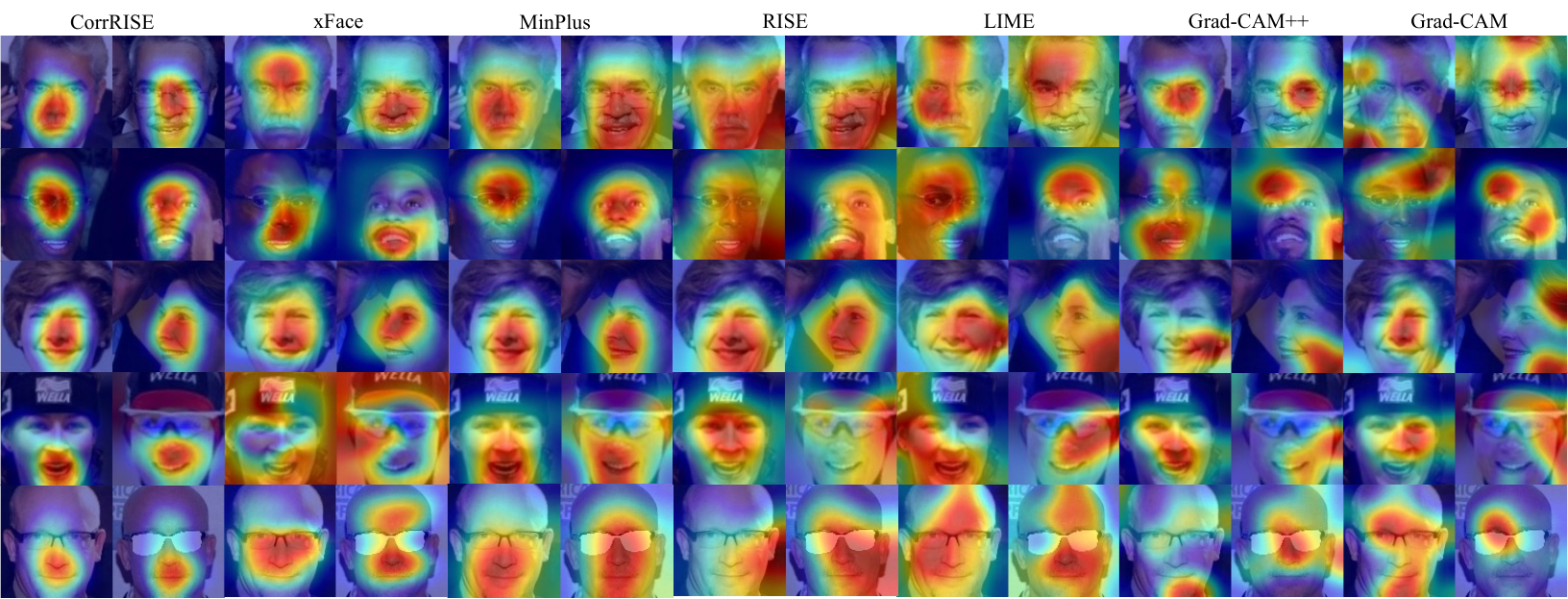}
	\end{adjustbox}
\caption{Visual results comparison among six saliency map-based explanation methods. The importance increases from blue to red color.}
\label{fig:comparison}
\end{figure*}

\subsubsection{Comparison with other Explanation Approaches}
A visual comparison is made between the proposed method and six explainable face verification approaches, including adaptations of four XAI approaches, i.e. Grad-CAM \cite{selvaraju2017grad}, Grad-CAM++ \cite{chattopadhay2018grad}, LIME \cite{ribeiro2016should}, and RISE \cite{petsiuk2018rise}, and two state-of-the-art XFR methods, namely MinPlus \cite{mery2022true} and xFace \cite{knoche2023explainable}. Both MinPlus and xFace consist of several variations and the best-performing one is selected. Implementation details refer to Section \ref{label:implementation}. 
Due to the limitation of several methods, only the similar regions between two images are visualized and compared. As a result, Fig.~\ref{fig:comparison} shows that CorrRISE consistently obtains more stable saliency maps than other approaches and can always precisely highlight the most similar regions between any given image pair. The adapted Grad-CAM and Grad-CAM++ methods produce less stable and meaningful explanation maps. LIME and RISE tend to allocate a very broad range of high-saliency pixels, making the importance map less precise. xFace and MinPlus achieve comparable results with CorrRISE. But the former fails in a difficult sample (row 4), and the latter does not exclude mask regions (row 5).

On the other hand, the samples in Fig. \ref{fig:verification} and \ref{fig:comparison} are selected from diverse demographic groups, e.g. various genders, ethnicities, and ages. The visual results have validated that the CorrRISE method only depends on the decision of the verification system and shows no significant bias across demographic groups. 




\begin{table}[t]
  \centering
  \caption{Quantitative evaluation of saliency maps using proposed Deletion and Insertion metrics (\%) on LFW, CPLFW, and Webface-OCC datasets, representing three different verification scenarios. \textbf{Del~($\downarrow$)} refers to the Deletion metric, the smaller the better. \textbf{Ins~($\uparrow$)} refers to the Insertion metric, the larger the better. }
    \resizebox{0.48\textwidth}{!}{
    \begin{tabular}{c|cc|cc|cc}
    \toprule
    \multirow{2}[4]{*}{Methods} & \multicolumn{2}{c|}{LFW} & \multicolumn{2}{c|}{CPLFW} & \multicolumn{2}{c}{WebfaceOcc} \\
\cmidrule{2-7}          & Del   & Ins   & Del   & Ins   & Del   & Ins \\
    \midrule
    Grad-CAM\cite{selvaraju2017grad} & 50.65 & 65.00 & 40.15 & 56.77 & 35.33 & 69.67 \\
    Grad-CAM++\cite{chattopadhay2018grad} & 46.40 & 68.02 & 37.08 & 58.26 & 36.78 & 68.51 \\
    LIME\cite{ribeiro2016should}  & 38.48 & 78.72 & 28.83 & 70.59 & 24.04 & 82.15 \\
    RISE\cite{petsiuk2018rise}  & 34.40 & 81.83 & 30.86 & 67.71 & 28.00  & 84.25   \\
    \midrule
    xFace\cite{knoche2023explainable} & 35.21 & 79.95 & 32.34 & 65.69 & 27.72 & 81.17 \\
    MinPlus\cite{mery2022true} & 29.98 & 85.24 & 21.27 & 75.24 & 18.32 & 89.22 \\
    \midrule
    CorrRISE & \textbf{23.29} & \textbf{85.70} & \textbf{17.60} & \textbf{77.88} & \textbf{13.24} & \textbf{89.70} \\
    \bottomrule
    \end{tabular}%
    }
  \label{tab:eval}%
\end{table}%

\subsection{Quantitative Evaluation Results}

This section reports the ``Deletion'' and ``Insertion'' metrics for a quantitative evaluation for general explainable face verification models. As described in Section \ref{chapter:eval}, the metrics measure the changes in verification accuracy after modifying the input images according to the saliency map. Considering that some methods cannot deal with non-matching cases and for a fair comparison, evaluation is conducted based on the matching pairs of each dataset. 



Table \ref{tab:eval} shows the quantitative comparison among the state-of-the-art explanation methods with the deletion and insertion metrics in three typical face verification scenarios. In general, the reported metrics are consistent with the visual observations in Fig. \ref{fig:comparison} and those visual results in supplementary materials. For example, the adapted Grad-CAM and Grad-CAM++ show poorer performance than other SOTA explanation methods. 
It is notable that CorrRISE achieves much better quantitative results than a straightforward adaptation of RISE. The latter can only obtain comparable results with other adapted XAI methods such as LIME, demonstrating the advantage and value of our proposed CorrRISE method. Although the two XFR methods, xFace and MinPlus, present visually compelling results, quantitative metrics show that CorrRISE is capable of providing more precise saliency maps for all three datasets. 



\begin{table}[t]
  \centering
  \caption{Explainablity performance of CorrRISE tested on different state-of-the-art face recognition models. The verification accuracy (\%) of FR models and two explainability metrics (\%) for CorrRISE are reported.}
    \resizebox{\linewidth}{!}{%
    \begin{tabular}{c|c|cc}
    \toprule
    FR Models & Acc (LFW) & Deletion ($\downarrow$) & Insertion ($\uparrow$) \\
    \midrule
    ArcFace \cite{deng2019arcface} & 99.53 & 23.29 & 85.70 \\
    MagFace \cite{meng2021magface} & 99.83 & 23.35 & 85.68  \\
    AdaFace \cite{kim2022adaface} & 99.82 & 23.31 & 85.63 \\
    \bottomrule
    \end{tabular}%
    }
  \label{tab:crossmodel}%
\end{table}%

CorrRISE is additionally applied to three different state-of-the-art FR models to show the validity and generalization ability of our proposed method. Table \ref{tab:crossmodel} shows that, when all the models achieve similar verification accuracy on LFW dataset, the generated saliency maps also report similar scores on ``Deletion'' and ``Insertion'' metrics, which proves that CorrRISE is model-agnostic.





\section{Conclusion}

In this work, the problem of explanation for face verification has been explored. Specifically, a new explanation framework for face verification was conceived. The proposed CorrRISE algorithm produces precise and insightful saliency maps to interpret the decision-making process of a deep FR model. Extensive experimental results in multiple scenarios show the advantage of our proposed method and its capability in analyzing the potential verification failures in challenging scenarios and, thereby provide insights into improving the current FR system. 
Meanwhile, a new evaluation methodology is designed, which offers a fair comparison among the saliency map-based explainable face verification methods and benefits future research in this area. 


{\small
\bibliographystyle{ieee_fullname}
\bibliography{egbib}
}

\end{document}